\documentclass[10pt,journal,compsoc]{IEEEtran}
\ifCLASSOPTIONcompsoc
  \usepackage[nocompress]{cite}
\else
  \usepackage{cite}
\fi
\usepackage{hyperref} % -- This package is specifically forbidden
\hypersetup{
    colorlinks=true,
    linkcolor=red,
}
\usepackage{amssymb}
\usepackage{bbding}
\usepackage{booktabs}

\ifCLASSINFOpdf
   \usepackage[pdftex]{graphicx}
\else
\fi

\usepackage{stfloats}
\usepackage{color}
\usepackage[dvipsnames]{xcolor}

\usepackage{multirow}

\usepackage{caption}
\usepackage{subcaption}
\usepackage{amsmath}
\usepackage{amssymb}
\usepackage{amsmath}
\usepackage{epsfig}
\usepackage{amssymb}
\usepackage{amsmath}
\usepackage{graphicx}
\usepackage{amsmath}
\usepackage{amssymb}
\usepackage{booktabs}
\usepackage{amsfonts} % or
\usepackage{tabularx}
\usepackage{wrapfig}
\usepackage{caption}
\usepackage{array}
\usepackage{amsthm}
\usepackage{graphics}
\usepackage{graphicx}
\usepackage{enumitem}
\usepackage{multirow}
\usepackage{wasysym}
\usepackage{soul}
\usepackage{svg}
\usepackage{tikz}
\usepackage{wasysym}
\usepackage{xspace}
\usepackage{siunitx}
\usepackage{mathtools}
\usepackage{algorithm}
\usepackage{algorithmicx}
\usepackage[noend]{algpseudocode}
\usepackage{subcaption}
\usepackage{xparse}
\usepackage{dsfont}
\usepackage{colortbl}
\usepackage{float}
\usepackage{bm}
\usepackage{pifont}
\usepackage{dashbox}
\usepackage{sourcecodepro}
\usepackage{changepage}
\usepackage{rotating} % For rotating the table
\usepackage{hyperref}

\usepackage{amssymb}% http://ctan.org/pkg/amssymb
\usepackage{pifont}% http://ctan.org/pkg/pifont
\newcommand{\ours}{\textbf{ALPHA}\xspace}

\begin{document}
\title{
Prototype-Guided Pseudo-Labeling with Neighborhood-Aware Consistency for Unsupervised Adaptation
}

\author{
    Eman~Ali$^{1,3}$,
    Chetan~Arora$^2$,
    and~Muhammad~Haris~Khan$^1$%
    
    \IEEEcompsocitemizethanks{
        \IEEEcompsocthanksitem $^1$ Mohamed Bin Zayed University of Artificial Intelligence, Abu Dhabi, UAE.
        \IEEEcompsocthanksitem $^2$ Indian Institute of Technology Delhi, New Delhi, India.
        \IEEEcompsocthanksitem $^3$ Alexandria University, Alexandria, Egypt.
    }
}

\markboth{Journal of \LaTeX\ Class Files}%
{Shell \MakeLowercase{\textit{et al.}}: Bare Demo of IEEEtran.cls for Computer Society Journals}

\IEEEtitleabstractindextext{%
\begin{abstract} 
In unsupervised adaptation for vision-language models such as CLIP, pseudo-labels derived from zero-shot predictions often exhibit significant noise, particularly under domain shifts or in visually complex scenarios. Conventional pseudo-label filtering approaches, which rely on fixed confidence thresholds, tend to be unreliable in fully unsupervised settings. In this work, we propose a novel adaptive pseudo-labeling framework that enhances CLIP's adaptation performance by integrating prototype consistency and neighborhood-based consistency. The proposed method comprises two key components: PICS, which assesses pseudo-label accuracy based on in-class feature compactness and cross-class feature separation; and NALR, which exploits semantic similarities among neighboring samples to refine pseudo-labels dynamically. Additionally, we introduce an adaptive weighting mechanism that adjusts the influence of pseudo-labeled samples during training according to their estimated correctness. Extensive experiments on 11 benchmark datasets demonstrate that our method achieves state-of-the-art performance in unsupervised adaptation scenarios, delivering more accurate pseudo-labels while maintaining computational efficiency.
\end{abstract}

}

\maketitle

\IEEEdisplaynontitleabstractindextext

% \fi
\IEEEpeerreviewmaketitle

\IEEEraisesectionheading{\section{Introduction}\label{sec:introduction}}

\label{sec:intro}
Unsupervised adaptation of vision-language models (VLMs), particularly CLIP~\cite{clip}, leverages its zero-shot classification capability to generate pseudo-labels for target-domain samples without requiring labeled data~\cite{upl, pouf}. However, these pseudo-labels are often unreliable under domain shifts or in complex settings like satellite imagery~\cite{clip, Wang2022DebiasedLF}, leading to error accumulation during training and reduced adaptation performance.

One potential strategy to address the challenge of noisy pseudo-labels in unsupervised CLIP adaptation is to adopt confidence-based filtering, as proposed in FixMatch~\cite{sohn2020fixmatch}, which filters out low-confidence predictions to reduce the propagation of incorrect pseudo-labels.
However, their effectiveness diminishes in fully unsupervised settings due to the lack of ground-truth supervision. 
Additionally, CLIP’s inherent issues—such as miscalibrated confidence scores and a bias toward imbalanced predictions~\cite{Wang2022DebiasedLF}—further hinder reliable pseudo-label selection and often amplify the effects of label noise.

To address these limitations, recent unsupervised adaptation methods have proposed more sophisticated self-training frameworks tailored to VLMs. LaFTer~\cite{lafter}, for example, utilizes large language models (LLMs) to generate instance-level descriptions and trains an auxiliary classifier to refine pseudo-labels. However, it does not incorporate explicit filtering mechanisms for the generated pseudo-labels, which may compromise robustness across diverse datasets.
In contrast, ReCLIP~\cite{reclip} enhances label reliability via cross-modal self-training and label propagation, selecting high-confidence pseudo-labels based on agreement between modalities. However, its inductive framework is computationally intensive, and dependence on label propagation hampers scalability, particularly for large-scale datasets. Consequently, the fundamental challenge persists: how to reliably refine pseudo-labels under domain shift without external supervision.

\begin{figure*}[t]
  \centering
  \includegraphics[width=\textwidth]{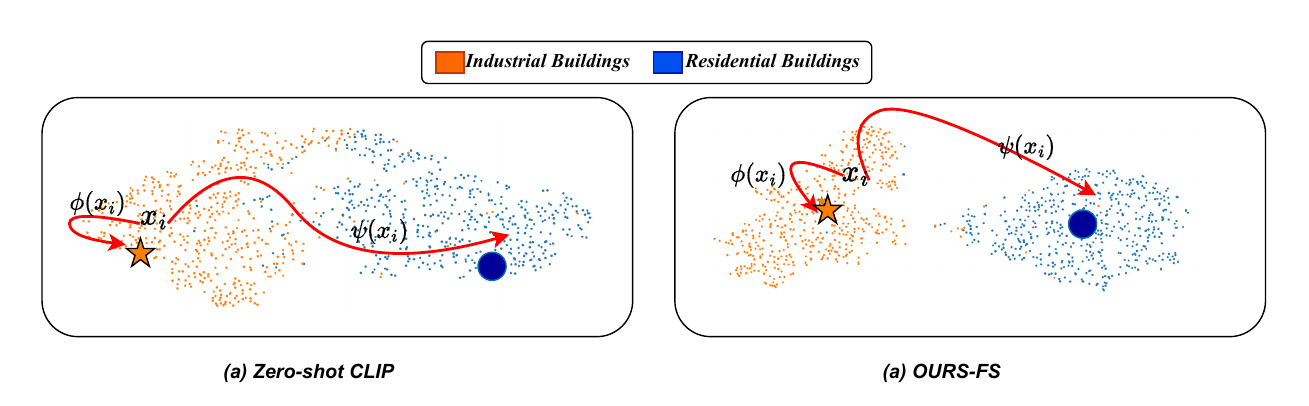}
  \caption{t-SNE visualization of visual embeddings for zero-shot CLIP~\cite{clip} and \ours-FS on EuroSAT, showing textual (circle $\newmoon$) and visual (star $\bigstar$) prototypes. Metrics include in-class similarity, $\phi(x_i)$, and cross-class separation, $\psi(x_i)$. Image $x_i$ is near its visual prototype and distant from the most confident samples of other classes.}
  \label{fig:motivate}
\end{figure*}

In this paper, we address the challenge of reliable pseudo-labeling in unsupervised CLIP adaptation with a robust and scalable framework. Although CLIP’s zero-shot capability is promising, it often struggles to maintain pseudo-label accuracy under distribution shifts. To overcome this, we propose \ours (\textbf{A}daptive Pseudo-\textbf{L}abeling via \textbf{P}rototype Consistency and neig\textbf{H}borhood \textbf{A}wareness), a method that enhances pseudo-label accuracy by leveraging both spatial and semantic structures in the embedding space.
Drawing inspiration from prototype-based techniques in low-supervision settings~\cite{Yang2023PrototypeGuidedPL}, our approach consists of two key components. First, \textbf{PICS} (\textbf{P}rototype-based \textbf{I}ntra-class and \textbf{C}ross-class \textbf{S}coring) evaluates in-class compactness and cross-class separation in feature space to filter out ambiguous samples lacking strong prototype alignment (see Figure~\ref{fig:motivate}). Second, \textbf{NALR} (\textbf{N}eighbor-guided \textbf{A}daptive \textbf{L}abel \textbf{R}efinement) refines pseudo-labels by exploiting local semantic coherence among neighbors. Additionally, we introduce a dynamic weighting scheme that emphasizes locally consistent samples, improving training stability, especially during the early stages.

Our contributions are twofold. First, we present a prototype and neighbor-aware framework for unsupervised CLIP adaptation, addressing the noisy pseudo-label through geometric and semantic structure. This includes \textbf{PICS}, which improves label reliability via in- and cross-class scoring. Second, we enhance CLIP's textual prototypes using LLM-generated descriptive class texts and introduce \textbf{NALR}, a neighborhood-driven label refinement module that reinforces local consistency. Extensive experiments on 11 challenging benchmarks confirm our method's superiority over state-of-the-art approaches, with strong generalization across diverse domains.

\section{Related works}
\label{sec:related_work}
\noindent \textbf{Unsupervised adaptation for CLIP:}  VLMs like CLIP~\cite{clip} have demonstrated impressive zero-shot visual recognition capabilities by leveraging large-scale image-text pretraining. By aligning visual and textual embeddings, CLIP enables robust predictions across diverse datasets. 
Recent efforts to improve CLIP’s adaptability have focused on fine-tuning additional parameters using few-shot samples such as CoOp~\cite{coop}, MaPLe~\cite{khattakMaPLe}, and Tip-Adapter~\cite{zhang2022tip}, which show promising results on challenging tasks. However, these approaches depend on labeled samples, which are often scarce or costly in many practical scenarios~\cite{huang2024adapting}.
To address this limitation, unsupervised adaptation techniques exploit unlabeled target data for fine-tuning. For example, UPL~\cite{upl} generates pseudo-labels from zero-shot CLIP predictions, while LaFTer~\cite{lafter} employs online adaptive pseudo-labeling by treating all generated labels as supervision for strongly augmented views. Confidence-based filtering methods, such as ReCLIP~\cite{reclip}, selectively train on high-confidence pseudo-labels to reduce noise.
Despite these advances, domain shifts often degrade CLIP’s prediction reliability, causing noisy pseudo-labels that propagate errors and impair adaptation performance~\cite{Wang2022DebiasedLF}.
In this work, we introduce a novel unsupervised adaptation framework that leverages the intrinsic structure of the embedding space, based on the assumption that clean samples exhibit strong intra-class compactness and clear inter-class separation. This enables accurate filtering of clean pseudo-labels from noisy ones.  

\noindent\textbf{Pseudo-labeling} is a fundamental technique widely used in semi-supervised and unsupervised learning to effectively utilize unlabeled data. For example, FixMatch~\cite{sohn2020fixmatch} introduced a simple yet effective confidence-thresholding method, while S2Match~\cite{guan2025s2match} proposed a dynamic sampling strategy that prioritizes high-confidence predictions to reduce the impact of noisy labels. To tackle class imbalance and enhance pseudo-label selection, FGBC~\cite{kong2023fgbc} employs a flexible graph-based balanced classifier with class-aware threshold adjustments.
Further improvements in pseudo-label reliability include ensemble methods like EPLE~\cite{wang2024evidential} that mitigate label noise, and PADCLIP~\cite{Wang2022DebiasedLF}, which explicitly focuses on debiasing pseudo-labels. Managing noisy pseudo-labels remains a critical challenge: some methods refine labels through semantic clustering to address severe noise~\cite{sachdeva2023scanmix}, while others discard unreliable labels to prevent error propagation~\cite{wang2022learning}.
Our approach complements these strategies by leveraging semantic information from text descriptions generated via LLMs to refine pseudo-labels. Additionally, we introduce an adaptive weighting scheme that assesses pseudo-label accuracy based on neighborhood consistency, enabling more nuanced sample weighting during training.

%% ============================================================ METHOD ============================================================ 
\section{Methodology}
\label{sec:methodology}

\begin{figure*}[!t]
    \centering
    \includegraphics[width=\textwidth]{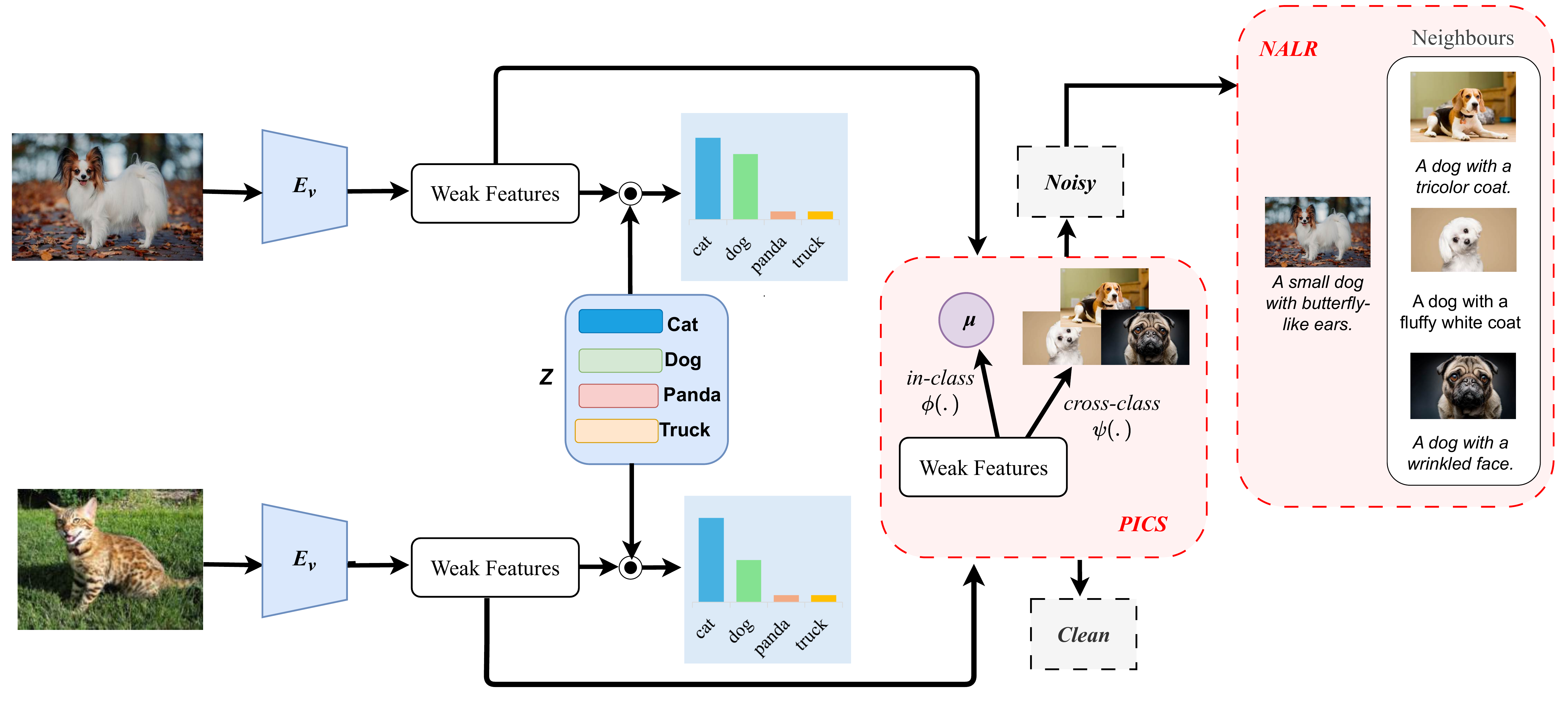} % Adjust width as needed
    \captionsetup{singlelinecheck=off}
    \caption{Overview of \ours, consisting of two components:  
    (a) \textbf{PICS}: Assesses pseudo-label accuracy via in-class and cross-class metrics. In-class compactness measures similarity between image features and their class prototype, $\mu$, while cross-class separation evaluates dissimilarity with other classes.  
    (b) \textbf{NALR}: Enhances noisy pseudo-labels by enforcing neighbor consistency, aligning predictions with neighboring samples.}
    \label{fig:main_architecture}
\end{figure*}

\subsection{Zero-shot visual classification in CLIP}
\label{ZS_clip}
The zero-shot CLIP model excels at generalizing across domains without task-specific training, delivering strong performance on unlabeled datasets. Given a pre-trained CLIP model \( (E_v, E_t) \), with \( E_v \) as the visual encoder and \( E_t \) as the textual encoder, we apply it to an unlabeled dataset \( \mathcal{D}_t = \{ x_i \}_{i=1}^N \), where \( x_i \in \mathcal{X}_t \) represents an image, and \( \mathcal{C} = \{ c_j \}_{j=1}^C \) denotes the set of class names.
For each class in \( \mathcal{C} \), the textual encoder \( E_t \) generates embeddings by processing hand-crafted prompts, producing a set of text prototypes \( \mathbf{Z} \in \mathbb{R}^{C \times d} \), where \( d \) is the embedding dimensionality. The zero-shot prediction for an image \( x_i \), denoted \( \hat{y}_i \), is determined by the class with the highest cosine similarity between the image’s visual embedding \( E_v(x_i) \) and the text prototypes \( \mathbf{Z} \):
\begin{equation}
    \hat{y}_i = \arg\max_{c \in \mathcal{C}} \zeta(E_v(x_i), \mathbf{Z}_c),
    \label{eq:zero-shot}
\end{equation}
where \( \zeta \) is the cosine similarity function, and \( \mathbf{Z}_c \) is the text prototype for class \( c \).

\subsection{Unsupervised adaptation of CLIP}
\label{UA_clip}
To adapt CLIP to the unlabeled target dataset \( \mathcal{D}_t \) in an unsupervised manner, we use a pseudo-labeling approach. This involves generating pseudo-labels for weakly augmented images, denoted \( \alpha(x_i) \), by computing the similarity between their visual features and textual prototypes \( \mathbf{Z} \), as follows:
\begin{equation}
    p(x_i) = \zeta(E_v(\alpha(x_i)), \mathbf{Z}),
    \label{eq:pseudo_text}
\end{equation}
Using the similarities computed in Eq.~\ref{eq:pseudo_text}, we assign a pseudo-label \( \hat{y}_i \) to each image by selecting the class with the highest similarity score:
\begin{equation}
    \hat{y}_i = \arg\max p(x_i).
\end{equation}
This pseudo-label \( \hat{y}_i \) supervises the training of a strongly augmented version of the image, \( \mathcal{A}(x_i) \). The supervised learning process is governed by the self-training loss:
\begin{equation}
    \mathcal{L}_{st} = -\frac{1}{N} \sum_{i=1}^N \log p(y = \hat{y}_i \mid E_v(\mathcal{A}(x_i))),
    \label{eq:loss_clip}
\end{equation}
where the conditional probability \( p(y = \hat{y}_i \mid E_v(\mathcal{A}(x_i))) \) is defined as:
\begin{equation}
    p(y = \hat{y}_i \mid E_v(\mathcal{A}(x_i))) = \frac{\exp \left( \zeta(E_v(\mathcal{A}(x_i)), \mathbf{Z}_{\hat{y}_i}) \right)}{\sum_{c=1}^C \exp \left( \zeta(E_v(\mathcal{A}(x_i)), \mathbf{Z}_c) \right)},
    \label{eq:likelihood_clip}
\end{equation}
To mitigate confirmation bias in CLIP~\cite{Wang2022DebiasedLF}, we incorporate a fairness regularization loss from~\cite{li2022masked} into the training process. This loss is defined as:
\begin{equation}
    \mathcal{L}_{reg} = -\frac{1}{C} \sum_{j=1}^C \log \bar{p}_{\mathcal{A}(x_j)},
\end{equation}
where \( \bar{p}_{\mathcal{A}(x_j)} \) is the batch-wise average predicted probability across classes for the strongly augmented image \( \mathcal{A}(x_j) \). The predicted probabilities \( p_{\mathcal{A}(x_j)} \) are derived from the similarity between the visual features \( E_v(\mathcal{A}(x_j)) \) and textual prototypes \( \mathbf{Z} \). This regularization promotes balanced adaptation by preventing overfitting to pseudo-labels, encouraging a uniform distribution of predictions across all \( C \) classes~\cite{li2022masked}.

\subsection{Adaptive pseudo-label filtering}
\label{clean_detection}
Unsupervised adaptation of CLIP using the self-training loss \( \mathcal{L}_{st} \) from Eq.~\ref{eq:loss_clip}, even with fairness regularization \( \mathcal{L}_{reg} \), can introduce confirmation bias. Pseudo-labels generated by zero-shot CLIP are often inaccurate, especially in complex domains~\cite{clip}. Traditional filtering methods that rely on high-confidence thresholds are ineffective in fully unsupervised settings due to the absence of ground-truth supervision~\cite{sohn2020fixmatch}.
To address these challenges, we propose \textbf{PICS} (\textbf{P}rototype-based \textbf{I}ntra-class and \textbf{C}ross-class \textbf{S}coring), a novel method for detecting noisy pseudo-labels. Unlike confidence-based approaches, \textbf{PICS} leverages in-class compactness and cross-class separation—core principles of representation learning—to distinguish accurate pseudo-labels from noisy ones without ground-truth supervision. 
As illustrated in Figure~\ref{fig:main_architecture}, \textbf{PICS} operates on the principle that a correct pseudo-label corresponds to an image with high similarity to its assigned class and low similarity to other classes. It assesses pseudo-label accuracy by analyzing the sample’s position in the embedding space using two key metrics: in-class compactness (how closely a sample aligns with its class prototypes) and cross-class separation (how distinct it is from other classes). This spatial analysis enables \textbf{PICS} to identify reliable pseudo-labels in a fully unsupervised manner effectively. 

\subsubsection{In-class compactness computation}
To measure in-class compactness, we construct class prototypes by aggregating image embeddings in a confidence-aware manner. Using image embeddings \( \{ f_i \}_{i=1}^N \), where \( f_i = E_v(x_i) \), we create prototypes for each of the \( C \) classes. In the absence of ground-truth labels, we leverage unlabeled data through a two-step process: (1) generate pseudo-labels \( \hat{y}_i \) for each image, and (2) compute confidence scores \( \omega_i = \max_{c \in C} p(x_i) \), derived from Eq.~\ref{eq:pseudo_text}. The class prototype for each class \( c \) is computed as a weighted average of embeddings, incorporating confidence scores:
\begin{equation}
    \mu_c = \frac{\sum_{i: \hat{y}_i = c} \omega_i f_i}{\sum_{i: \hat{y}_i = c} \omega_i}, \quad c = 1, 2, \ldots, C,
    \label{eq:proto}
\end{equation}
where the summation is over images assigned to class \( c \), and \( \mu_c \) represents the prototype for class \( c \).
To improve prototype quality and reduce the impact of noisy pseudo-labels, we employ a memory bank mechanism~\cite{Zhou2023SourcefreeDA} that stores feature representations, pseudo-labels, and confidence scores for each image. The memory bank is defined as:
\begin{equation}
    \mathbf{MB} = \{(f_i, \hat{y}_i, \omega_i)\}_{i=1}^N,
\end{equation}
Throughout training, \( \mathbf{MB} \) is continuously updated to reflect the latest representations. At the end of each epoch, class prototypes are recomputed using the stored features, pseudo-labels, and confidence scores, enabling iterative refinement of prototypes. 
The in-class compactness for each image is measured by the cosine similarity between its feature representation \( f_i \) and the prototype of its assigned class \( \mu_{\hat{y}_i} \), defined as the in-class score:
\begin{equation}
    \phi(x_i) = \zeta(f_i, \mu_{\hat{y}_i}),
    \label{eq:intra_prot}
\end{equation}
where \( \phi(x_i) \) quantifies how closely the image aligns with its class prototype.

\subsubsection{Cross-class separation computation}
\label{sec:out_class_selection}
In-class compactness measures how closely an image aligns with its assigned class prototypes, while cross-class separation quantifies its dissimilarity from features of other classes. To compute cross-class separation, we construct a cross-class set \( \mathcal{O} \) from the memory bank \( \mathbf{MB} \), comprising samples from classes different from the image's assigned class. 
We propose three distinct methods for constructing \( \mathcal{O} \), each tailored to select representative samples: Confidence-based top-\( k \) selection (\textbf{CS}), Random sampling (\textbf{RS}), and Confusion-based top-\( k \) selection (\textbf{FS}). These methods capture diverse aspects of the data distribution, ensuring robust and effective sample selection for cross-class separation. The details of each method are described below.

\noindent \textbf{\textbf{C}onfidence-based top-$k$ \textbf{S}election (CS):}
The \textbf{CS} method prioritizes samples from non-matching classes with the highest confidence scores to focus on the most reliable representations. For each image feature \( f_i \), we retrieve from the memory bank \( \mathbf{MB} \) all samples with pseudo-labels differing from \( \hat{y}_i \). These samples are ranked by their confidence scores, and the cross-class set \( \mathcal{O}(f_i) \) is formed by selecting the top-\( k \) most confident samples, as follows:
\begin{equation}
    \mathcal{O}(f_i) = \operatorname{Top-}k_{\omega} \left\{ f_j \mid \hat{y}_j \neq \hat{y}_i, (f_j, \hat{y}_j, \omega_j) \in \mathbf{MB} \right\},
    \label{eq:confident}
\end{equation}
where \( \operatorname{Top-}k_{\omega} \) denotes selecting the \( k \) samples with the highest confidence weights \( \omega_j \). 

\noindent \textbf{\textbf{R}andom \textbf{S}ampling (RS):} method constructs the cross-class set for image features \( f_i \) by randomly selecting \( k \) samples from the memory bank \( \mathbf{MB} \) with pseudo-labels different from \( \hat{y}_i \), as follows:
\begin{equation}
    \mathcal{O}(f_i) = \operatorname{Random-}k \left\{ f_j \mid \hat{y}_j \neq \hat{y}_i, (f_j, \hat{y}_j, \omega_j) \in \mathbf{MB} \right\},
    \label{eq:random}
\end{equation}
where \( \operatorname{Random-}k \) denotes the random selection of \( k \) feature vectors \( f_j \).

\begin{figure}[!t]
  \centering
  \includegraphics[width=\linewidth]{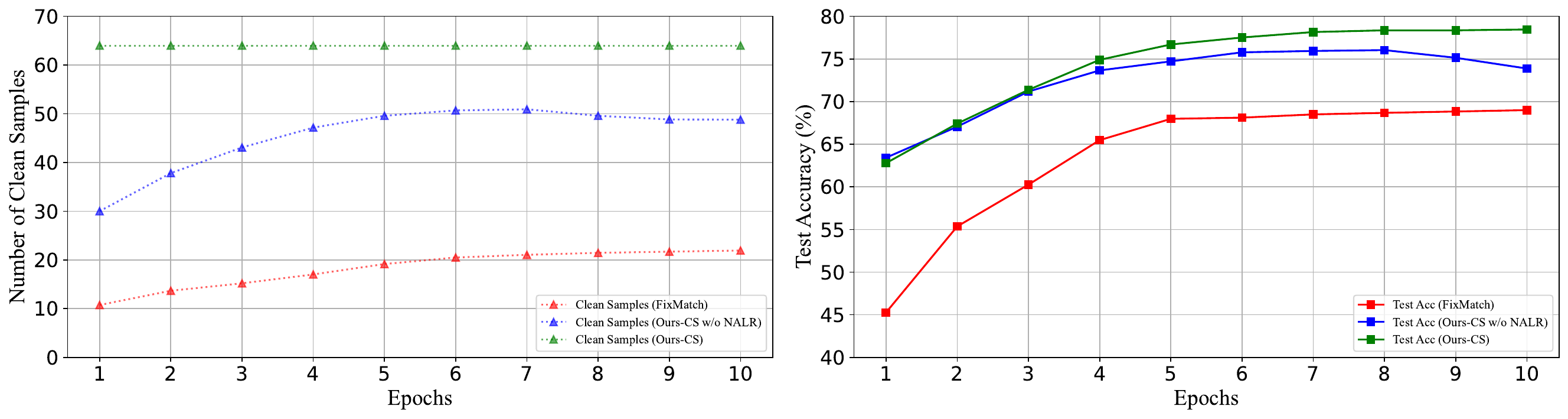}
  \caption{Comparison of clean sample counts (left) and test accuracy (right) for FixMatch-based threshold~\cite{sohn2020fixmatch} (threshold=0.95), \ours-CS w/o \textbf{NALR}, and \ours-CS, evaluated on the EuroSAT dataset.}
  \label{fig:fix_match}
\end{figure}

\noindent \textbf{Confusion-based top-\( k \) Selection (FS):}  
The \textbf{FS} method identifies samples likely to cause classification confusion by focusing on the confused class. For each image feature \( f_i \), the pseudo-label \( \hat{y}_i \) is assigned based on the highest similarity score, while the class with the second-highest score, denoted \( \hat{y}_i^{(2)} \), represents the confused class, computed as:
\begin{equation}
    \hat{y}_i^{(2)} = \arg\max_{c \in C \setminus \{\hat{y}_i\}} \zeta(E_v(x_i), \mathbf{Z}),
\end{equation}
From the memory bank \( \mathbf{MB} \), we collect samples with pseudo-labels matching \( \hat{y}_i^{(2)} \). The cross-class set \( \mathcal{O}(f_i) \) is then formed by selecting the top-\( k \) most confident samples, as follows:
\begin{equation}
    \mathcal{O}(f_i) = \operatorname{Top-}k_{\omega} \left\{ f_j \mid \hat{y}_j = \hat{y}_i^{(2)}, (f_j, \hat{y}_j, \omega_j) \in \mathbf{MB} \right\},
    \label{eq:confused}
\end{equation}
After constructing the cross-class set \( \mathcal{O}(f_i) \) using one of the methods described above, the cross-class separation \( \psi(x_i) \) is computed as the average similarity between its image features \( f_i \) and the features in \( \mathcal{O}(f_i) \):
\begin{equation}
    \psi(x_i) = \frac{1}{|\mathcal{O}(f_i)|} \sum_{f' \in \mathcal{O}(f_i)} \zeta(f_i, f'),
    \label{eq:out_class}
\end{equation}
The self-training loss \( \mathcal{L}_{st} \) is applied only to samples with clean pseudo-labels, identified by comparing their in-class compactness \( \phi(x_i) \) with their cross-class separation \( \psi(x_i) \). A pseudo-label is deemed clean if \( \phi(x_i) > \psi(x_i) \), indicating that the sample is more similar to its own class than to other classes. The self-training loss is defined as:
\begin{equation}
    \mathcal{L}_{st} = -\frac{1}{N} \sum_{i=1}^N \mathbb{I}\{\phi(x_i) > \psi(x_i)\} \log p(y = \hat{y}_i \mid E_v(\mathcal{A}(x_i))),
    \label{eq:loss_clean}
\end{equation}
where \( \mathbb{I}\{\phi(x_i) > \psi(x_i)\} \) is an indicator function that equals 1 if \( \phi(x_i) > \psi(x_i) \), and 0 otherwise.This method refines pseudo-labels by prioritizing clean samples, allowing the model to focus on accurate pseudo-labels during training, thereby improving performance, as shown in Figure~\ref{fig:fix_match}.

\subsection{Leveraging neighborhood consistency for noisy label refinement}
\label{noisy_correct}
Although \textbf{PICS} outperforms a naive high-confidence filter method~\cite{sohn2020fixmatch}, as shown in Figure~\ref{fig:fix_match}, the availability of clean samples is often limited at the beginning of training due to CLIP's inherent miscalibration and its tendency to produce inaccurate predictions~\cite{Wang2022DebiasedLF}, resulting in a shortage of accurate pseudo-labels. 
This can lead to overfitting on these samples, while a portion of the data remains unutilized during training. To address this, we propose \textbf{NALR}—\textbf{N}eighbor-guided \textbf{A}daptive \textbf{L}abel \textbf{R}efinement, a method to refine and leverage these ignored noisy pseudo-labels more effectively. 

\noindent Inspired by the work in cross-modal matching~\cite{NACN}, which handles noisy correspondence by leveraging the semantic consistency of neighboring samples, we extend it to the unsupervised adaptation setting. We introduce \textbf{NALR} to rectify noisy pseudo-labels by incorporating neighborhood relationships, assigning cleaner pseudo-labels to the noisy ones, thereby enabling more effective noise handling. 
However, we do not have direct image-text correspondences since our problem setting differs from cross-modal matching. We leverage CuPL~\cite{CuPL} to generate ensembles of class-specific descriptions using LLMs to address this issue. For each class $c$, an LLM, $h(.)$, generates multiple descriptive texts using the prompt: ``What does a \{$c$\} look like?'' This process yields a set of textual descriptions for each class label in the target dataset: 
\begin{equation}
    h(c) = \{ t^c_j \}_{j=1}^{M}, \quad
    \mathbf{T} = \{ h(c) \}_{c \in C}
\end{equation}
where $M$ denotes the total number of generated descriptions per class, and $\mathbf{T}$ represents the complete set of descriptions for all classes.
To assign pseudo-labels, we first compute the similarity between the image features $f_i$ and the embeddings of all text descriptions in $\mathbf{T}$, as shown in \eqref{eq:select_text}. The most similar text description is then mapped to its corresponding class label by applying the text-to-label mapping function $g$, as shown in \eqref{eq:map_label}.
\begin{equation} \label{eq:select_text}
    \hat{r}_i = \arg\max_{t \in \mathbf{T}} \zeta(f_{i}, E_t(\mathbf{T}))
\end{equation}
\begin{equation} \label{eq:map_label}
    \hat{y}^{h}_i = g(t_{\hat{r}_i})
\end{equation}
Where $\hat{r}_i$ is the index of the most similar text description in $\mathbf{T}$, $t_{\hat{r}_i}$ is the text description corresponding to $\hat{r}_i$, and $g(t_{\hat{r}_i})$ maps the selected text description to its corresponding class label. 
Formally, we redefine the training dataset for the noisy pseudo-labels by including both the pseudo-labels and their corresponding text descriptions as follows:
\begin{equation}
    \mathcal{D}_t = \{(x_i, \hat{y}^h_i, t_{\hat{r}_i})\}_{i=1}^N,
\end{equation}
This approach ensures that each image with a noisy pseudo-label is assigned a more accurate label based on its alignment with more sophisticated text descriptions, using semantic information generated by LLMs. However, these descriptions and their corresponding pseudo-labels may still be noisy due to potential mismatches between the generated descriptions and the global image view~\cite{wca}.
To address this, we propose a novel \textit{adaptive weighting mechanism} based on the premise that, in a clean dataset, semantically similar texts are associated with visually similar images~\cite{NACN}. 
For each textual description \(t_{\hat{r}_i}\) we retrieve its top-$k$ most similar textual neighbors, denoted as \(\{ t_{\hat{r}_{i,j}} \}_{j=1}^{k_n}\). Similarly, for the image-text pair \( (x_i, t_{\hat{r}_i}) \), we locate its top-$k$ nearest neighbors, represented as \( \{(x_{i,j}, t_{\hat{r}_{i,j}})\}_{j=1}^{k_n} \). We denote this set of neighbors as \( \mathcal{N}_i = \{(x_{i,j}, t_{\hat{r}_{i,j}})\}_{j=1}^{k_n} \). Building on that, we introduce an adaptive weighting mechanism to assess the accuracy of each image-text pair \( (x_i, t_{\hat{r}_i}) \). This mechanism assigns a weight to each image-text pair based on its similarity relative to its neighbors, computed as:
\begin{equation}
    \lambda_i = \sigma(\Delta \zeta_i),
\end{equation}
where \( \sigma(\cdot) \) is the sigmoid function and \( \Delta \zeta_i \) is the similarity difference, defined as:
\begin{equation}
    \Delta \zeta_i = \zeta(f_i, E_t(t_{\hat{r}_i})) - \frac{1}{k_n} \sum_{j=1}^{k_n} \zeta(f_{i,j}, E_t(t_{\hat{r}_{i,j}})).
\end{equation}
where \( \zeta(f_i, E_t(t_{\hat{r}_i})) \) is the similarity between the image features \( f_i \) and the embedding of the text description \( t_{\hat{r}_i} \), and the second term is the average similarity of the \( k_n \) nearest neighbor pairs \( (x_{i,j}, t_{\hat{r}_{i,j}}) \). 
The sigmoid function constrains the weight \( \lambda_i \) to the range \( [0,1] \). This mechanism adjusts each sample's weight based on its estimated accuracy: a negative \( \Delta \zeta_i \) indicates that the image-text pair \( (x_i, t_{\hat{r}_i}) \) is likely noisy, resulting in a lower confidence weight, while a positive \( \Delta \zeta_i \) suggests higher accuracy, yielding a greater weight. 
This adaptive weighting enables \textbf{NALR} to dynamically modulate the influence of each sample according to its accuracy, enhancing robustness against noisy pseudo-labels. To further improve performance, we update the memory bank with refined pseudo-labels generated by this method, as described in Eq.~\ref{eq:memory_update}. This integration establishes a feedback loop that continuously refines the model's accuracy and resilience to noise.
\begin{equation}
    \label{eq:memory_update}
    \mathbf{MB} =
    \begin{cases} 
        \{(f_i, \hat{y}_i, \omega_i)\}_{i=1}^N, & \text{if } \phi(x_i) > \psi(x_i), \\ 
        \{(f_i, \hat{y}^h_i, \lambda_i)\}_{i=1}^N, & \text{otherwise},
    \end{cases}
\end{equation}
The condition \( \phi(x_i) > \psi(x_i) \) selects reliable pseudo-labels \( \hat{y}_i \) with confidence \( \omega_i \); otherwise, text-derived pseudo-labels \( \hat{y}^h_i \) are used with \( \omega_i = \lambda_i \). 
The loss function for noisy pseudo-labels is defined as:
\begin{equation}
    \label{eq:loss_proto}
    \mathcal{L}_N = -\frac{1}{N} \sum_{i=1}^N \mathbb{I}\{\phi(x_i) \leq \psi(x_i)\} \lambda_i \log p(y = \hat{y}^h_i \mid E_v(\mathcal{A}(x_i))),
\end{equation}
Finally, the overall loss function used for self-training is $\mathcal{L} = \mathcal{L}_{st} + \mathcal{L}_{N} + \mathcal{L}_{reg}$.

%% ============================================================ EXPERIMENTS ============================================================ 

\section{Experiments}
\label{sec:framework}

\noindent\textbf{Datasets and baselines:} We perform an extensive evaluation of our approach, \ours across 11 diverse datasets: Caltech101~\cite{fei2004learning}, DTD~\cite{cimpoi2014describing}, EuroSAT~\cite{helber2019eurosat}, Flowers102~\cite{nilsback2008automated}, OxfordPets~\cite{parkhi2012cats}, UCF101~\cite{soomro2012ucf101}, StanfordCars~\cite{krause20133d}, \sloppy Food101~\cite{bossard2014food}, CIFAR100~\cite{krizhevsky2009learning}, CUB-200-2011~\cite{cub}, and RESISC45~\cite{Cheng2017RemoteSI}. These datasets cover a wide range of domains, enabling a robust assessment of generalization.
We benchmark our method against the most recent and state-of-the-art (SOTA) unsupervised adaptation methods for CLIP. These approaches represent the forefront of unsupervised adaptation and zero-shot learning for CLIP, ensuring a comprehensive and competitive evaluation: 
\begin{itemize} 
    \item \textbf{CLIP}~\cite{clip}: A robust zero-shot image classification model leveraging contrastive vision-language pretraining.
    \item \textbf{CuPL}~\cite{CuPL}: Enhances zero-shot classification by initializing classifiers with averaged class-specific text descriptions generated by LLMs.
    \item \textbf{UPL}~\cite{upl}: Fine-tunes CLIP with learnable textual prompts in the text encoder, trained using top-$k$ per-class confidence sampling from zero-shot CLIP predictions.
    \item \textbf{POUF}~\cite{pouf}: Fine-tunes prompts on unlabeled target data by aligning discrete distributions from text prompts as class prototypes with target image features.
    \item \textbf{LaFTer}~\cite{lafter}: Utilizes GPT-3-generated text descriptions to train a text classifier, producing pseudo-labels for effective CLIP fine-tuning.
    % \item \textbf{ReCLIP}~\cite{reclip}: Learns a projection space to correct misaligned visual-text embeddings, generates pseudo-labels, and uses cross-modality self-training to iteratively refine encoders, reducing domain gap and misalignment.
    \item \textbf{DPA}~\cite{Ali_2025_WACV}: Generates accurate pseudo-labels by combining and ranking outputs from two prototypes to reduce noise early in training, while aligning textual and image prototypes to address visual-textual misalignment.
\end{itemize}

\noindent For \ours, we report results from three proposed strategies for selecting the cross-class set, as described in Section~\ref{sec:out_class_selection}. These are denoted as \ours-CS (Eq.~\ref{eq:confident}), using confidence-based top-$k$ selection; \ours-RS (Eq.~\ref{eq:random}), employing random sampling; and \ours-FS (Eq.~\ref{eq:confused}), utilizing confusion-based top-$k$ selection.

\noindent\textbf{Implementation details:} 
We employ CLIP ViT-B/32, pre-trained by OpenAI~\cite{clip}. During unsupervised adaptation, we fine-tune only the layer normalization parameters of the image encoder and the text prototypes to maintain stability under noisy supervision~\cite{reclip}. Drawing inspiration from CuPL~\cite{CuPL}, we initialize the text prototypes \( \mathbf{Z} \) by averaging $M$ text descriptions per class generated by LLMs, following the same number of text descriptions as in CuPL, and discard the text encoder post-initialization.
For data augmentation, we apply strong transformations, including RandomResizedCrop, Flip, and RandAugment. Pseudo-label generation uses Resize and RandomCrop augmentations, while inference employs CenterCrop. The model is optimized with the AdamW optimizer, using a cosine learning rate schedule, a batch size of 64, and training for 15 epochs. A learning rate of $5 \times 10^{-5}$ is used for all datasets, except StanfordCars and Food101, which use $1 \times 10^{-6}$.
For \textbf{PICS} and \textbf{NALR}, we set $k=3$ and $k_n=3$, respectively. All experiments are conducted on a single NVIDIA Quadro RTX 6000 GPU. 
To ensure fair comparisons, we reproduce SOTA results using their publicly available codebases and standardize dataset splits with VISSL~\cite{goyal2021vissl}, resolving inconsistencies in data splits across SOTA methods.

%% -------------------------------------------- Main Results --------------------------------------------------
\subsection{Comparisons with state-of-the-art}
Table~\ref{tab:sota_vit_b_32} presents the results of \ours, showcasing substantial improvements across multiple benchmarks. Our analysis demonstrates that all variants of our method (\ours-CS, \ours-RS, and \ours-FS) consistently outperform SOTA approaches. The robust performance across diverse cross-class set selection strategies highlights the effectiveness and versatility of our proposed framework.
Notably, \ours-CS achieves significant gains over Zero-shot CLIP and CuPL, with average accuracy improvements of $+10.07\%$ and $+7.85\%$, respectively, across all datasets. Compared to UPL, which emphasizes prompt optimization, \ours-CS delivers an average gain of $+10.28\%$. UPL tends to underperform Zero-shot CLIP when evaluated on the full unlabeled training dataset, likely due to unreliable predictions from Zero-shot CLIP~\cite{lafter}. Additionally, \ours-CS outperforms POUF, LaFTer, and DPA, achieving average accuracy improvements of $+10.14\%$, $+7.46\%$, and $+1.42\%$, respectively, across the 11 evaluated datasets.

\begin{table*}
    \centering
    \resizebox{\textwidth}{!}{
    \begin{tabular}{@{}lccccccccccccc@{}}
      \toprule
      \textbf{Method} & \textbf{Venue} & \textbf{Caltech} & \textbf{DTD} & \textbf{ESAT} & \textbf{Flower} & \textbf{OxPets} & \textbf{UCF} & \textbf{Cars} & \textbf{Food} & \textbf{CIFAR100} & \textbf{CUB} & \textbf{RES45} & \textbf{Avg} \\
      \midrule
      \rowcolor[gray]{0.9} \multicolumn{14}{c}{\textbf{Zero-shot Methods}} \\
      \midrule
      \textbf{CLIP}~\cite{clip} & ICML’21 & 90.69 & 44.42 & 43.84 & 66.46 & 87.46 & 64.24 & 58.74 & 82.40 & 65.10 & 51.76 & 57.59 & 64.79 \\
      \textbf{CuPL}~\cite{CuPL} & ICCV’23 & 93.92 & 50.11 & 50.06 & 68.05 & 87.16 & 66.96 & 60.80 & 84.03 & 65.20 & 49.71 & 61.14 & 67.01 \\
      \midrule
      \rowcolor[gray]{0.9} \multicolumn{14}{c}{\textbf{UA Methods}} \\
      \midrule
      \textbf{UPL}~\cite{upl} & - & 92.36 & 45.37 & 51.88 & 67.00 & 83.84 & 62.04 & 49.41 & 84.25 & 67.41 & 49.22 & 57.63 & 64.58 \\
      \textbf{POUF}~\cite{pouf} & ICML’23 & 94.10 & 46.10 & 62.90 & 40.00 & 87.80 & 61.20 & 57.70 & 82.10 & 62.00 & 51.59 & 66.40 & 64.72 \\
      \textbf{LaFTer}~\cite{lafter} & NeurIPS’23 & 94.39 & 50.32 & 69.96 & 67.80 & 84.93 & 65.08 & 57.44 & 82.45 & 69.79 & 37.66 & 61.60 & 67.40 \\
      % \textbf{ReCLIP}~\cite{reclip} & WACV’24 & 93.84 & 53.88 & 69.48 & 72.63 & 87.11 & 66.67 & 58.84 & 84.22 & 71.43 & 53.95 & 73.05 & 71.37 \\
      
      \textbf{DPA}~\cite{Ali_2025_WACV} & WACV’25 & \textbf{95.94}&	55.96&	\textbf{79.94}	&75.56	&\underline{90.11}	&66.69	&56.83	&\underline{84.76}&	74.22	&\textbf{56.70}	&71.11&	73.44 \\
      \midrule
      \rowcolor[gray]{0.9} \multicolumn{14}{c}{\textbf{\ours{} (UA Method)}} \\
      \midrule
      \textbf{\ours-CS} & - & {94.94} & 57.93 & 78.18 & \textbf{76.78} & 90.00 & \textbf{72.61} & 60.81 & \textbf{84.80} & \underline{75.13} & \underline{56.54} & 75.71 & \textbf{74.86} \\
      
      \textbf{\ours-RS} & - & 94.34 & \textbf{58.67} & 78.18 & \underline{76.29} & 89.92 & \underline{72.46} & \underline{61.12} & 84.73 & \textbf{75.14} & 56.06 & \underline{75.83} & \underline{74.79} \\
      
      \textbf{\ours-FS} & - & \underline{95.02} & \underline{57.98} & \underline{78.48} & 76.09 & \textbf{90.22} & 72.27 & \textbf{61.15} & 84.65 & 74.65 & 55.85 & \textbf{75.97} & 74.76 \\
      \bottomrule
    \end{tabular}
    }
    \caption{Accuracy (\%) of \ours compared to SOTA methods across 11 diverse datasets using the ViT-B/32 backbone.}
    \label{tab:sota_vit_b_32}
\end{table*} 

\subsection{Ablation study}
For the ablation study, we select 6 of the 11 datasets, excluding larger ones—StanfordCars, Food101, CIFAR-100, CUB-200-2011, and RESISC45—to ensure computational efficiency. This focused selection allows extensive experimentation within resource constraints.

%% -------------------------------------------- Components analysis ----------------------------------------------
\subsubsection{Components analysis}
To evaluate the contributions of \ours's components, we train four distinct model variants:

\begin{itemize}
    \item \ours-B: Fine-tunes CLIP using the loss function from Eq.~\ref{eq:loss_clip} + $\mathcal{L}_{reg}$, without \textbf{PICS} or \textbf{NALR}. All generated pseudo-labels are used for self-training without filtering or refinement.
    
    \item \ours w/o \textbf{NALR}: Fine-tunes CLIP using the loss function from Eq.~\ref{eq:loss_clean} + $\mathcal{L}_{reg}$, leveraging \textbf{PICS} to select clean samples while excluding noisy samples (i.e., omitting \textbf{NALR}). We explore three sub-variants based on cross-class set selection: \ours-CS w/o \textbf{NALR}, \ours-RS w/o \textbf{NALR}, and \ours-FS w/o \textbf{NALR}.
    
    \item \ours w/o $\lambda$: Fine-tunes CLIP using the loss function from Eq.~\ref{eq:loss_clean} + Eq.~\ref{eq:loss_proto} + $\mathcal{L}_{reg}$, but omits the adaptive weight $\lambda$ in \textbf{NALR}. We assess this variant across cross-class set selection strategies: \ours-CS w/o $\lambda$, \ours-RS w/o $\lambda$, and \ours-FS w/o $\lambda$, to evaluate the role of adaptive weighting.
    
    \item \ours w/o \textbf{PICS}: Fine-tunes CLIP using the loss function from Eq.~\ref{eq:loss_proto} + $\mathcal{L}_{reg}$, incorporating \textbf{NALR} but excluding \textbf{PICS}. No pseudo-label filtering is applied in this configuration.
\end{itemize}

\noindent As presented in Table~\ref{tab:ablation}, \ours-B significantly outperforms the zero-shot CLIP baseline, primarily due to the inclusion of $\mathcal{L}_{reg}$, which effectively reduces confirmation bias during training.
By incorporating \textbf{PICS} for pseudo-label filtering, \ours w/o \textbf{NALR} further enhances performance, demonstrating its capability to identify and mitigate noisy pseudo-labels without requiring label refinement. However, slight performance reductions are observed in specific cases: \ours-CS w/o \textbf{NALR} on OxfordPets and \ours-FS w/o \textbf{NALR} on Caltech101. These reductions likely stem from overly aggressive filtering by \textbf{PICS}, which may discard some informative samples in datasets with high intra-class variability, leading to a loss of discriminative information.
Adding \textbf{NALR} without adaptive weighting (\ours w/o $\lambda$) results in a marginal performance drop compared to \ours w/o \textbf{NALR}. This is attributed to occasional mismatches between generated text descriptions and the global image context~\cite{wca}, underscoring the critical role of the adaptive weight $\lambda$ in balancing label refinement within \ours.
In contrast, \ours w/o \textbf{PICS} achieves a modest 1.75\% improvement over \ours-B, highlighting \textbf{NALR}'s contribution to performance gains. However, performance decreases on datasets like DTD and OxfordPets are observed, likely due to \textbf{NALR}'s reliance on unfiltered pseudo-labels, which can propagate noise in datasets with complex textures or fine-grained class distinctions.
The complete \ours model, combining \textbf{PICS} and \textbf{NALR}, delivers the highest performance. This synergy leverages \textbf{PICS} for robust pseudo-label filtering and \textbf{NALR} for adaptive label refinement, yielding a highly accurate and resilient model across diverse datasets.

\begin{figure}[!ht]
  \centering
  \includegraphics[width=0.9\linewidth]{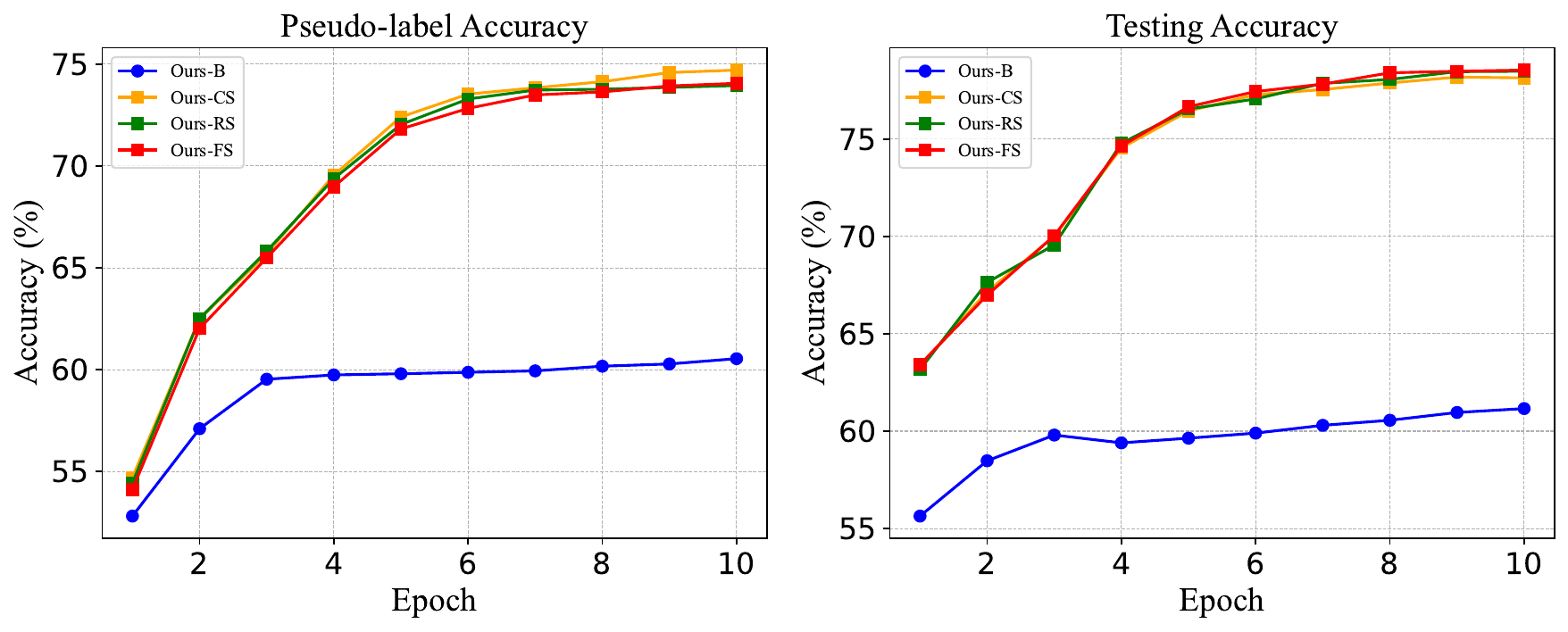}
  \caption{Comparison of pseudo-label accuracy (left) and test accuracy (right) on the EuroSAT dataset during training. The proposed methods (\ours-CS, \ours-FS, and \ours-RS) exhibit greater robustness to confirmation bias than the \ours-B baseline.}
  \label{fig:ours_vs_base}
\end{figure}

\noindent We provide a detailed analysis of the progression of pseudo-label accuracy and test accuracy across training epochs for \ours (\ours-CS, \ours-FS, and \ours-RS), compared against the baseline \ours-B model, as depicted in Figure~\ref{fig:ours_vs_base}. The results illustrate that \ours consistently sustains higher pseudo-label accuracy throughout training. This sustained accuracy highlights its robust capability to mitigate noisy pseudo-labels, thereby significantly enhancing overall model performance.

\begin{table}[!t]
  \centering
  \footnotesize
  \setlength{\tabcolsep}{3.5pt}
  \scalebox{0.85}[0.85]{
    \begin{tabular}{@{}lrrrrrrr@{}}
      \toprule
      \textbf{Method} & \textbf{Caltech} & \textbf{DTD} & \textbf{ESAT} & \textbf{Flower} & \textbf{OxPets} & \textbf{UCF} & \textbf{Avg} \\
      \midrule
      \rowcolor[gray]{0.9} \multicolumn{8}{c}{\textbf{Zero-shot Methods}} \\
      \midrule
      \textbf{Zero-shot CLIP}~\cite{clip} & 90.69 & 44.42 & 43.84 & 66.46 & 87.46 & 64.24 & 66.19 \\
      \textbf{CuPL}~\cite{CuPL} & 93.92 & 50.11 & 50.06 & 68.05 & 87.16 & 66.96 & 69.38 \\
      \midrule
      \rowcolor[gray]{0.9} \multicolumn{8}{c}{\ours{} (UA Method)} \\
      \midrule
      \ours-B & 93.48 & 54.04 & 61.12 & 71.78 & 90.30 & 68.41 & 73.19 \\
      \midrule
      \ours-CS w/o \textbf{NALR} & 93.89 & 57.98 & 73.86 & 75.56 & 89.94 & 71.42 & 77.11 \\
      \ours-RS w/o \textbf{NALR} & 94.01 & 58.30 & 76.70 & 75.56 & 90.46 & 71.24 & 77.71 \\
      \ours-FS w/o \textbf{NALR} & 93.42 & 58.14 & 76.74 & 75.07 & 90.41 & 70.98 & 77.46 \\
      \midrule
      \ours w/o \textbf{PICS} & 94.06 & 53.83 & 65.92 & 75.44 & 89.67 & 70.71 & 74.94 \\
      \midrule
      \textbf{\ours}-CS w/o $\lambda$ & 93.56 & 53.72 & 66.12 & 73.85 & 89.78 & 71.40 & 74.74 \\
      \textbf{\ours}-RS w/o $\lambda$ & 93.51 & 54.41 & 65.92 & 74.30 & 89.72 & 70.76 & 74.77 \\
      \textbf{\ours}-FS w/o $\lambda$ & 93.72 & 54.31 & 67.32 & 74.83 & 89.72 & 71.69 & 75.27 \\
      \midrule
      \ours-CS & 94.94 & 57.93 & 78.18 & 76.78 & 90.00 & 72.61 & 78.41 \\
      \ours-RS & 94.34 & 58.67 & 78.18 & 76.29 & 89.92 & 72.46 & 78.31 \\
      \ours-FS & \textbf{95.02} & 57.98 & \textbf{78.48} & 76.09 & \textbf{90.22} & \textbf{72.27} & \textbf{78.34} \\
      \bottomrule
    \end{tabular}
  }
  \caption{Ablation study on \ours components. \textbf{PICS} denotes \textbf{P}rototype-based \textbf{I}ntra-class and \textbf{C}ross-class \textbf{S}coring; \textbf{NALR} represents \textbf{N}eighbor-guided \textbf{A}daptive \textbf{L}abel \textbf{R}efinement. \textbf{CS}, \textbf{RS}, and \textbf{FS} indicate cross-class set selection strategies: \textbf{C}onfidence-based top-$k$ \textbf{S}election, \textbf{R}andom \textbf{S}ampling, and \textbf{F}usion-based top-$k$ \textbf{S}election, respectively. \ours-B is a simplified baseline model.}
  \label{tab:ablation}
\end{table}

\subsubsection{Scalability evaluation}
\begin{figure}[!ht]
  \centering
  \includegraphics[width=0.8\linewidth]{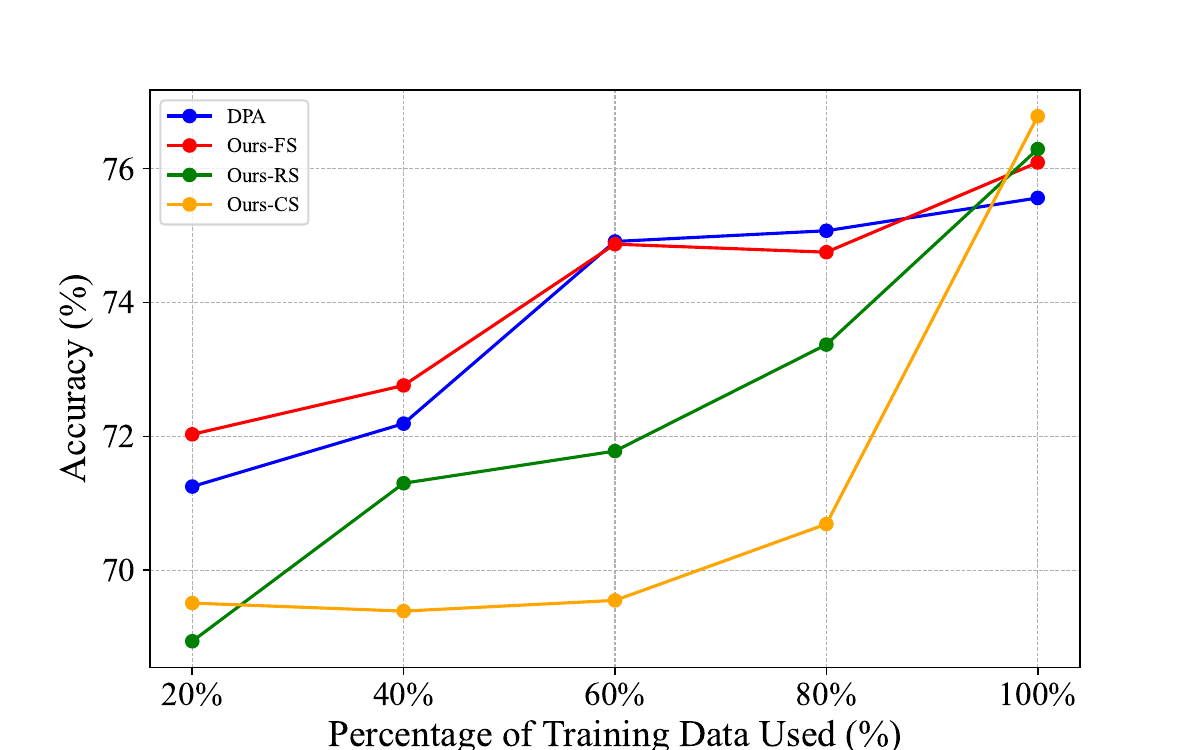}
  \caption{Scalability analysis of \ours and DPA on the Flowers dataset, showing accuracy across dataset subsets ($20\%, 40\%, 60\%, 80\%$, and $100\%$).}
  \label{fig:sacle}
\end{figure}

We evaluate the scalability of the proposed method, \ours, on the Flowers dataset by comparing its performance with DPA~\cite{Ali_2025_WACV} across varying proportions of the training data. Specifically, experiments are conducted using subsets comprising 20\%, 40\%, 60\%, and 80\% of the full training set. As shown in Figure~\ref{fig:sacle}, \ours consistently outperforms DPA when trained on the entire dataset. Both methods, however, exhibit a decline in performance as the training data size decreases, indicating sensitivity to reduced sample availability. Remarkably, the \ours-FS variant demonstrates strong robustness, achieving superior accuracy even when trained with only 20\% of the dataset, thereby highlighting its effectiveness in limited data scenarios.

\subsubsection{The impact of backbone networks}
To assess the performance of \ours using a different image encoder, we reduce the batch size from 64 to 32 to improve memory efficiency, relative to previous experiments conducted with ViT-B/32. The results obtained with ViT-B/16 are shown in Table~\ref{tab:sota_vit_b_16}. Our method consistently outperforms state-of-the-art approaches, highlighting its capability to enhance CLIP’s performance on downstream tasks when employing image encoders with finer patch granularity. These results demonstrate that \ours effectively leverages unlabeled data to improve CLIP’s robustness and generalization as model complexity increases.

\begin{table}[!t]
  \centering
  \footnotesize
  \setlength{\tabcolsep}{4pt}
  \scalebox{0.85}[0.85]{
    \begin{tabular}{@{}lccccccc@{}}
      \toprule
      \textbf{Method} & \textbf{Caltech} & \textbf{DTD} & \textbf{ESAT} & \textbf{Flower} & \textbf{OxPets} & \textbf{UCF} & \textbf{Avg} \\ 
      \midrule
      \rowcolor[gray]{0.9} \multicolumn{8}{c}{\textbf{Zero-shot Methods}} \\
      \midrule
      \textbf{Zero-shot CLIP}~\cite{clip} & 92.60 & 44.70 & 49.00 & 70.89 & 89.00 & 69.10 & 69.22 \\
      \textbf{CuPL}~\cite{CuPL} & 95.06 & 54.31 & 58.70 & 73.93 & 91.17 & 70.45 & 73.94 \\
      \midrule
      \rowcolor[gray]{0.9} \multicolumn{8}{c}{\textbf{UA Methods}} \\
      \midrule
      \textbf{UPL}~\cite{upl}      & 95.14 & 45.90 & 55.36 & 73.93 & 87.98 & 67.43 & 70.96 \\ 
      \textbf{POUF}~\cite{pouf}     & 95.40 & 48.60 & 59.50 & 72.10 & 91.80 & 71.50 & 73.15 \\ 
      \textbf{LaFTer}~\cite{lafter} & \underline{95.92} & 54.79 & 72.10 & 75.15 & 85.28 & 67.20 & 75.07 \\ 
      \textbf{DPA}~\cite{Ali_2025_WACV} & \textbf{96.09} & 50.32 & \textbf{81.22} & 78.64 & \textbf{93.35} & 74.44 & 79.01 \\
      \midrule
      \midrule
      \textbf{\ours-CS} & 94.88 & \underline{58.03} & 77.16 & \textbf{81.36} & 92.10 & 75.57 & \underline{79.85} \\ 
      \textbf{\ours-RS} & 94.85 & \textbf{58.09} & 73.32 & \underline{80.71} & \underline{92.59} & \underline{76.05} & 79.27 \\
      \textbf{\ours-FS} & 95.08 & 56.81 & \underline{80.88} & 80.35 & 92.37 & \textbf{76.24} & \textbf{80.29} \\ 
      \bottomrule
    \end{tabular}
  }
  \caption{Accuracy (\%) comparison of \ours against SOTA methods using CLIP-ViT-B/16.}
  \label{tab:sota_vit_b_16}
\end{table}

\subsubsection{Transductive setting}
We adopt the transductive setting following the protocol established by ReCLIP~\cite{reclip}, where model adaptation is performed directly on the unlabeled test set of each dataset prior to evaluation on the same set. Among current state-of-the-art unsupervised adaptation methods for VLMs, we select ReCLIP~\cite{reclip} and DPA~\cite{Ali_2025_WACV} as representative transductive approaches. ReCLIP addresses the misalignment between CLIP’s visual and textual embeddings by learning a projection subspace that suppresses redundant and class-agnostic features. It further enhances pseudo-label accuracy through label propagation~\cite{Iscen2019LabelPF} and cross-modal self-training on high-confidence predictions. To ensure a fair comparison, we modify ReCLIP by replacing its original handcrafted prompts with LLM-generated descriptions from CuPL~\cite{CuPL}; we refer to this variant as ReCLIP\textsuperscript{*} in Table~\ref{tab:reclip}. In contrast, DPA~\cite{Ali_2025_WACV} initializes text prototypes using a fixed prompt set and does not incorporate additional filtering mechanisms. 
Although ReCLIP is specifically tailored for transductive adaptation, we evaluate our method (\ours) against both ReCLIP and DPA under transductive and inductive settings to thoroughly assess its generalization and robustness. As shown in Table~\ref{tab:reclip}, \ours consistently outperforms both baselines across all settings, demonstrating strong adaptability and superior performance.

\begin{figure}[!t]
    \centering
    \includegraphics[width=0.8\linewidth]{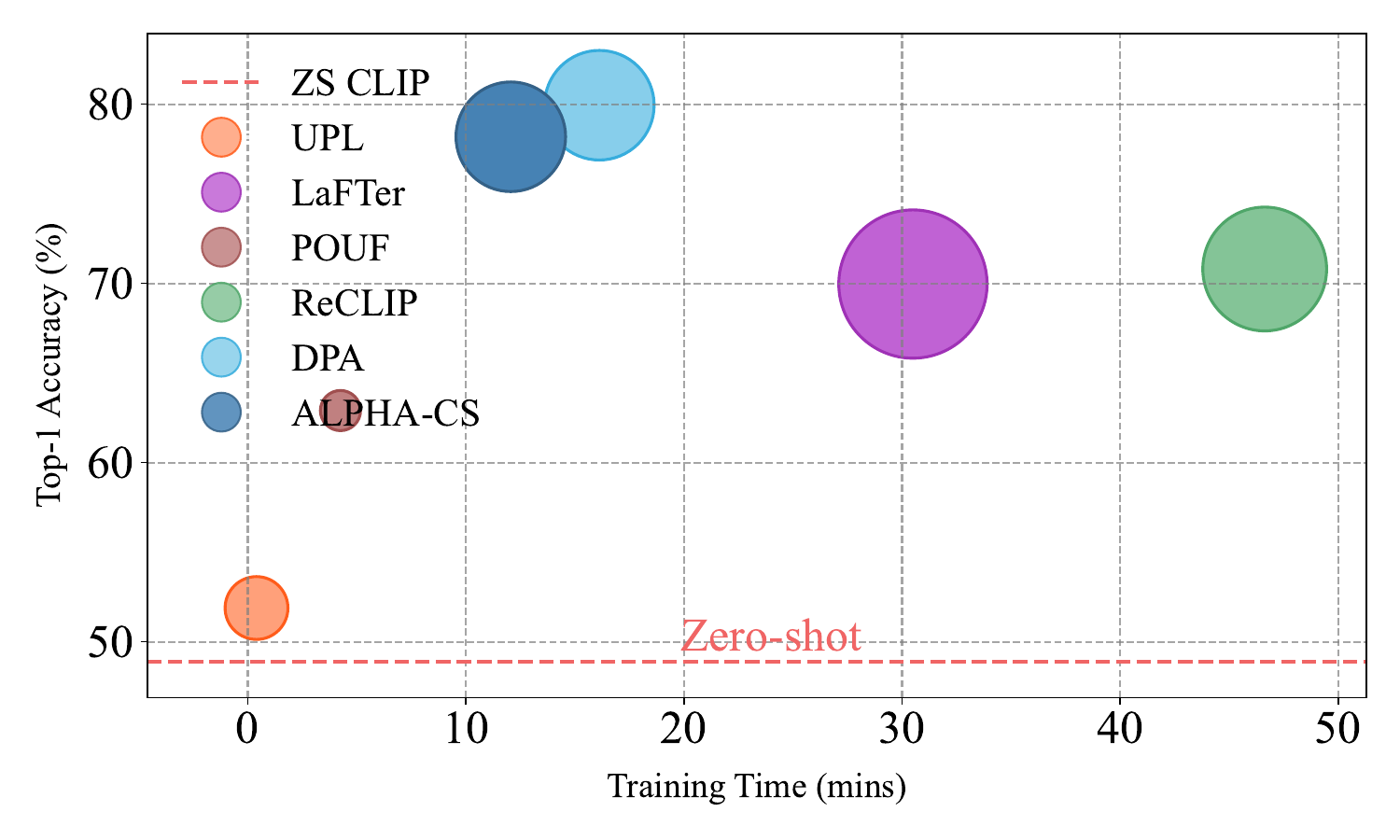}
     \caption{Efficiency comparison of \ours-CS and SOTA methods on the EuroSAT dataset. Circle radius indicates the number of trainable parameters for each method.}
    \label{fig:run}
\end{figure}

Additionally, we conduct a comparative study on the EuroSAT dataset to evaluate the computational efficiency of \ours-CS. As shown in Figure~\ref{fig:run}, \ours-CS achieves performance comparable to DPA, with only a slight decrease in accuracy but notable improvements in runtime efficiency. Moreover, it surpasses other state-of-the-art baselines while using fewer parameters and lower computational resources, highlighting its practicality for large-scale and resource-constrained applications.

\begin{table}[!ht]
  \centering
  \small
  \setlength{\tabcolsep}{5pt}
  \scalebox{0.85}[0.85]{
    \begin{tabular}{@{}lccccccc@{}}
      \toprule
      \textbf{Method} & \textbf{Caltech} & \textbf{DTD} & \textbf{ESAT} & \textbf{Flower} & \textbf{OxPets} & \textbf{UCF} & \textbf{Avg} \\ 
      \midrule
      \rowcolor[gray]{0.9} \multicolumn{8}{c}{\textbf{Inductive}} \\
      \midrule
          \textbf{ReCLIP} & {95.94} &53.88 &70.80 &72.63 &87.49 &67.01 &74.63 \\
          \textbf{ReCLIP}\textsuperscript{*} & 94.79 & 55.64&	70.64	&75.88&	\textbf{90.35}	&70.64& 76.32\\
          
          \textbf{DPA} & \textbf{95.94}	&55.96&	\textbf{79.94}&	75.56	&{90.11}	&66.69&	77.37\\
          
          \midrule
          \ours-CS &{94.94} &57.93 &{78.18} &\textbf{76.78} &90.00 &\textbf{72.61} &\textbf{78.41}\\
          
          \ours-RS &94.34 & \textbf{58.67} &{78.18} &\underline{76.29} &89.92 &\underline{72.46} &78.31\\
          
         \ours-FS &\underline{95.02} &\underline{57.98} &\underline{78.48} &76.09 &\underline{90.22} &72.27 &\underline{78.34}\\

      \midrule
      \rowcolor[gray]{0.9} \multicolumn{8}{c}{\textbf{Transductive}} \\
      \midrule
      \textbf{ReCLIP} &92.43 &52.50 &59.30 &70.65 &88.42 &69.13 &72.07\\
     \textbf{ReCLIP}\textsuperscript{*} &93.74	&53.19	&67.48	&74.99	&\underline{90.24}	&70.26	&74.98\\
     
     \textbf{DPA} & 91.62	&\textbf{58.09}	&\textbf{71.84}&	74.14	&\textbf{91.06}&	68.73	&75.91\\
      \midrule
          \ours-CS &\underline{94.58} &\underline{55.00} &71.46 &\underline{75.11} &89.64 &\underline{72.01} &76.30\\
          
          \ours-RS &\textbf{94.70} & {54.79} &{71.68} &\textbf{75.15} &89.72 &71.79 &\underline{76.31}\\
          
         \ours-FS &94.28 &54.63 &\underline{71.82} &75.03 &{89.78} &\textbf{72.46} &\textbf{76.33}\\
      \bottomrule
    \end{tabular}
  }
  \caption{Accuracy (\%) comparison of \ours, ReCLIP, and DPA across inductive and transductive settings.}
  \label{tab:reclip}
\end{table}

\subsubsection{The impact of fine-tuning strategies}
Our method, \ours, fine-tunes only the layer normalization (LN) parameters within CLIP’s image encoder, along with the learnable text prototypes $\mathbf{Z}$, while keeping all other parameters fixed. To evaluate the effectiveness of this selective tuning strategy, we compare \ours against several parameter-efficient fine-tuning methods applied to CLIP across multiple benchmark datasets. The comparative results are reported in Table~\ref{tab:full}. Specifically, we consider the following approaches:
\begin{itemize}
    \item \textit{LoRA}~\cite{hu2022lora}: Introduces trainable low-rank matrices into the model’s weight matrices to enable efficient adaptation with minimal parameter overhead.
    \item \textit{K-Adaptation}~\cite{he2023parameter}: Decomposes weight updates using Kronecker products; slow weights are shared across layers, and fast weights are further parameterized by low-rank matrices.
    \item \textit{Layer Normalization (LN)}~\cite{ba2016layer}: Restricts fine-tuning to only the LN parameters of the model.
\end{itemize}
As shown in Table~\ref{tab:full}, LN-only fine-tuning yields consistently strong performance across all datasets, with the variants \ours-CS, \ours-RS, and \ours-FS achieving competitive or superior accuracy. The results indicate that LN tuning performs on par with more complex methods such as K-Adaptation, highlighting both the robustness and the efficiency of \ours under limited trainable parameter budgets.

\begin{table}[!t]
  \centering
  \footnotesize
  \setlength{\tabcolsep}{4pt}
  \scalebox{0.85}[0.85]{
    \begin{tabular}{@{}lccccccc@{}}
      \toprule
      \textbf{Method} & \textbf{Caltech} & \textbf{DTD} & \textbf{ESAT} & \textbf{Flower} & \textbf{OxPets} & \textbf{UCF101} & \textbf{Avg} \\
      \midrule
      \rowcolor[gray]{0.9} \multicolumn{8}{c}{\ours{} (LoRA~\cite{hu2022lora})} \\
      \midrule
      \ours-CS & 92.11 & 53.03 & 54.48 & 75.48 & 88.53 & 69.76 & 72.23 \\
      \ours-RS & 91.08 & 54.84 & 48.76 & 75.19 & 89.75 & 69.68 & 71.55 \\
      \ours-FS & 91.57 & 54.89 & 44.90 & 74.30 & 88.63 & 69.23 & 70.59 \\
      \midrule
      \rowcolor[gray]{0.9} \multicolumn{8}{c}{\ours{} (K-Adaptation~\cite{he2023parameter})} \\
      \midrule
      \ours-CS & 94.70 & 58.09 & 76.80 & 76.90 & 89.92 & 72.03 & 78.07 \\
      \ours-RS & 94.79 & 57.93 & 76.96 & 76.70 & 90.19 & 72.35 & 78.15 \\
      \ours-FS & 94.96 & 58.30 & 76.96 & 76.17 & 90.16 & 72.38 & 78.15 \\
      \midrule
      \rowcolor[gray]{0.9} \multicolumn{8}{c}{\ours{} (LN)} \\
      \midrule
      \ours-CS & 94.94 & 57.93 & 78.18 & 76.78 & 90.00 & 72.61 & 78.41 \\
      \ours-RS & 94.34 & 58.67 & 78.18 & 76.29 & 89.92 & 72.46 & 78.31 \\
      \ours-FS & 95.02 & 57.98 & 78.48 & 76.09 & 90.22 & 72.27 & 78.34 \\
      \bottomrule
    \end{tabular}
  }
  \caption{Accuracy (\%) comparison of CLIP fine-tuning methods on \ours, using ViT-B/32 as the backbone for all experiments.}
  \label{tab:full}
\end{table}

\subsubsection{Robustness to noisy data}
\begin{figure}[!t]
  \centering
  \includegraphics[width=0.8\linewidth]{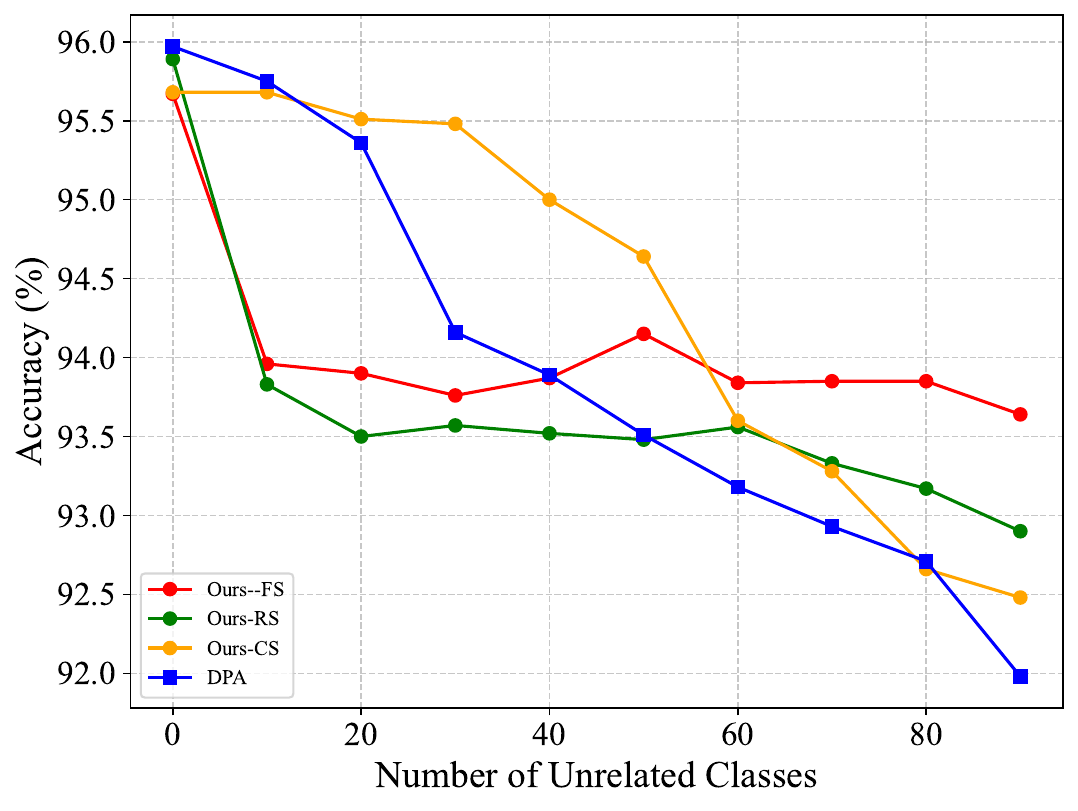}
  \caption{Accuracy (\%) comparison on CIFAR-10 with added unlabeled samples from unrelated CIFAR-100 classes, restricting target classes to CIFAR-10.}
  \label{fig:cifar10_noise}
\end{figure}

To evaluate the robustness of \ours under noisy conditions, we simulate label-irrelevant domain noise by injecting unlabeled images from unrelated CIFAR-100 classes into the CIFAR-10 training set, while restricting the target classes exclusively to CIFAR-10. The experimental setup is depicted in Figure~\ref{fig:cifar10_noise}. As anticipated, all methods experience performance degradation as the noise proportion increases. However, results in Figure~\ref{fig:cifar10_noise} show that \ours-CS consistently outperforms DPA~\cite{Ali_2025_WACV} across noise levels from 0\% up to 70\%. Moreover, \ours-RS and \ours-FS demonstrate greater stability at higher noise rates (50\%–90\%), with \ours-FS achieving the best overall performance under extreme noise conditions. These findings highlight the superior resilience of \ours variants against domain-irrelevant noisy samples.

\subsubsection{Sensitivity analysis}
\label{sec:lr_sense}
\begin{figure}[!t]
  \centering
  \includegraphics[width=\linewidth]{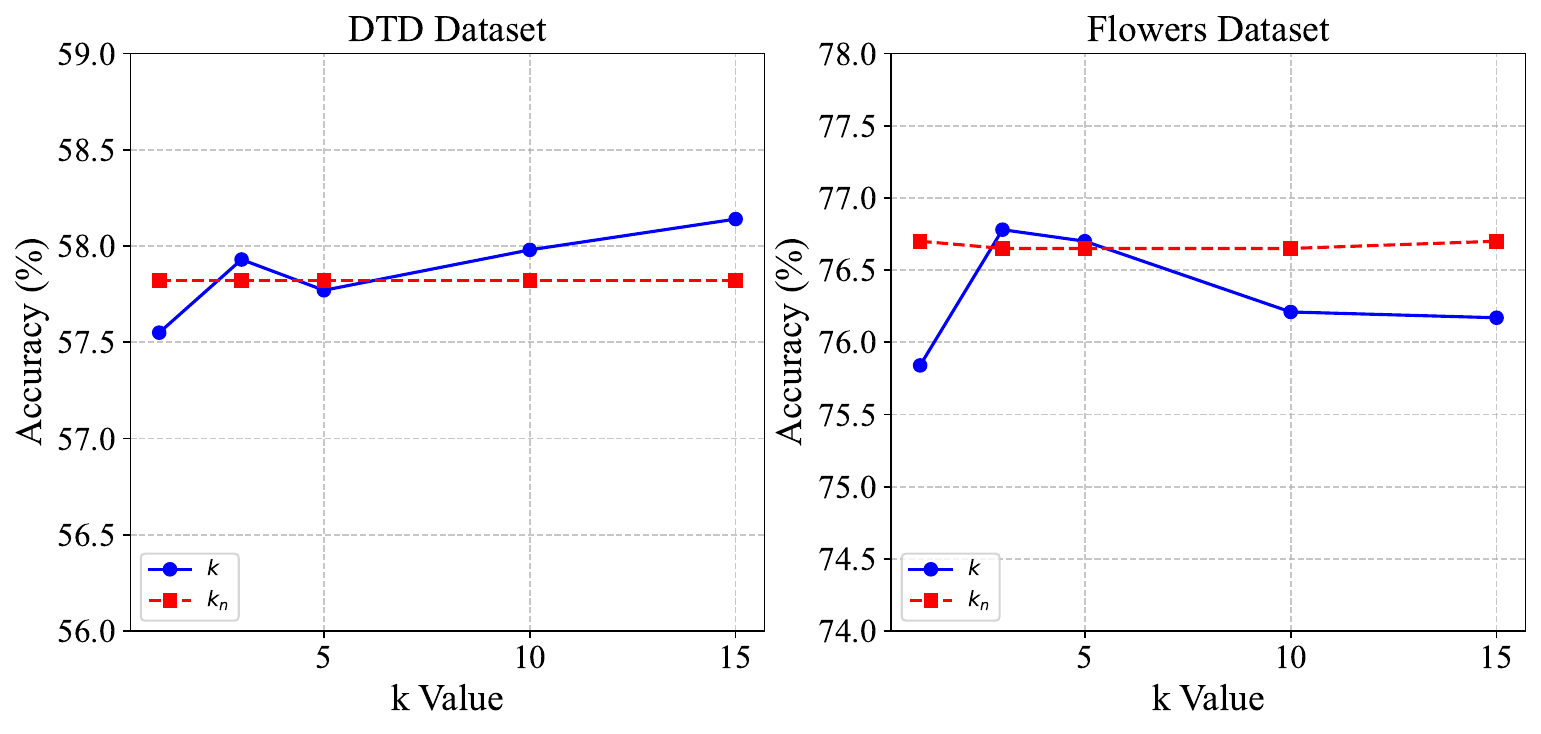}
  \caption{Impact of varying $k$ and $k_n$ on classification accuracy (\%) for \ours-CS on the DTD and Flowers datasets.}
  \label{fig:k_sense}
\end{figure}

We examine the sensitivity of \ours-CS to the hyperparameters $k_n$ and $k$, with results presented in Figure~\ref{fig:k_sense} for the DTD and Flowers datasets. On DTD, model accuracy increases steadily as $k$ ranges from 1 to 15, while variations in $k_n$ yield minimal impact, suggesting that performance on this dataset is predominantly influenced by $k$. Conversely, on the Flowers dataset, both $k$ and $k_n$ exhibit a substantial effect on accuracy, indicating a more complex interaction between these parameters in this setting.

%% ============================================================ conclusion ============================================================ 

\section{Conclusion}
\label{sec:conclusion}
In this paper, we introduce \ours, a novel framework designed to enhance CLIP’s performance in unsupervised adaptation, particularly in the presence of noisy pseudo-labels. By leveraging both spatial and semantic consistency within the embedding space, \ours effectively identifies and refines noisy pseudo-labels, thereby improving model robustness and overall accuracy. Our adaptive weighting mechanism further strengthens this process by dynamically adjusting the influence of pseudo-labels during training, enabling more reliable learning. 
Extensive experiments demonstrate that \ours consistently outperforms state-of-the-art methods, establishing it as a robust solution for unsupervised CLIP adaptation. However, our approach has limitations: the \textbf{PICS} module’s aggressive selectivity may inadvertently filter out informative samples in datasets with high intra-class variability, potentially hindering the model’s ability to capture subtle class distinctions.
Looking ahead, we aim to extend our framework beyond classification to other core visual recognition tasks such as image segmentation and object detection, where label noise remains a significant challenge. We believe these extensions will further showcase the versatility and impact of our approach across diverse and demanding applications.

\ifCLASSOPTIONcaptionsoff
  \newpage
\fi



% \bibliographystyle{IEEEtran}
% \bibliography{IEEEabrv,../bib/paper}
% \end{thebibliography}

{\small
\bibliographystyle{revision_ref}
\bibliography{ref}
}

\end{document}